%% file: main.tex
\documentclass[twoside]{article}

\usepackage[accepted]{aistats2025}
\usepackage{natbib}

\setcitestyle{numbers,sort,round}
\setcitestyle{authoryear,open={(},close={)}, aysep={,}} 
\usepackage{cleveref}
\usepackage{xcolor}
\usepackage{algorithm}
\usepackage{algpseudocode}
\usepackage{booktabs}       
\usepackage{graphicx} 
\usepackage{caption}
\usepackage{url}
\usepackage{rotating}
\usepackage{nicefrac}       
\usepackage{microtype}
\usepackage{bm}
\usepackage{amsmath}
\usepackage{amsthm}
\theoremstyle{definition}

\usepackage{pdfpages}

\usepackage{subcaption}

\usepackage{amssymb}
\usepackage{pifont}

\newcommand{\eg}{e.\,g., }
\newcommand{\ie}{i.\,e., }

\begin{document}

%

%
\runningauthor{Nakis, Kosma, Nikolentzos, Chatzianastasis, Evdaimon, Vazirgiannis}

\twocolumn[

\aistatstitle{Signed Graph Autoencoder for Explainable and Polarization-Aware Network Embeddings}





\aistatsauthor{ Nikolaos Nakis$^\dagger$ \And Chrysoula Kosma$^\ddagger$ \And Giannis Nikolentzos$^\spadesuit$}
\aistatsauthor{ Michail Chatzianastasis$^\dagger$ \And Iakovos Evdaimon$^\dagger$ \And Michalis Vazirgiannis$^\dagger$$^\clubsuit$}

\aistatsaddress{ $^\dagger$LIX, \'Ecole Polytechnique, Institute Polytechnique de Paris, France \\ $^\ddagger$Université Paris Saclay, Université Paris Cité, ENS Paris Saclay, CNRS, SSA, INSERM, Centre Borelli, France \\ $^\spadesuit$Department of Informatics and Telecommunications, University of Peloponnese, Greece \\ $^\clubsuit$Mohamed bin Zayed University of Artificial Intelligence, United Arab Emirates } ]

\begin{abstract}
  Autoencoders based on Graph Neural Networks (GNNs) have garnered significant attention in recent years for their ability to learn informative latent representations of complex topologies, such as graphs. Despite the prevalence of Graph Autoencoders, there has been limited focus on developing and evaluating explainable neural-based graph generative models specifically designed for signed networks. To address this gap, we propose the Signed Graph Archetypal Autoencoder (\textsc{SGAAE}) framework. \textsc{SGAAE} extracts node-level representations that express node memberships over distinct extreme profiles, referred to as archetypes, within the network. This is achieved by projecting the graph onto a learned polytope, which governs its polarization. The framework employs the Skellam distribution for analyzing signed networks combined with relational archetypal analysis and GNNs. Our experimental evaluation demonstrates the \textsc{SGAAE}'s capability to successfully infer node memberships over underlying latent structures while extracting competing communities.
Additionally, we introduce the \textsc{2-level} network polarization problem and show how \textsc{SGAAE} is able to characterize such a setting. 
The proposed model achieves high performance in different tasks of signed link prediction across four real-world datasets, outperforming several baseline models. Finally, \textsc{SGAAE} allows for interpretable visualizations in the polytope space, revealing the distinct aspects of the network, as well as, how nodes are expressing them. (\textit{Code available at}: \url{https://github.com/Nicknakis/SGAAE}).

\end{abstract}



\input{1-introduction.tex}

\input{2-related_work.tex}
\input{3-methods.tex}

\input{4-experiments}
\input{5-conclusion.tex}

\section*{Acknowledgements}
We gratefully acknowledge the reviewers for their constructive feedback and insightful comments. This work was supported by the French National research agency via the AML-HELAS (ANR-19-CHIA-0020) project.
M. C. is also supported by the EUR BERTIP (ANR-18-EURE-0002), Plan France 2030.
C. K. is supported by the IdAML Chair hosted at ENS Paris-Saclay, Université Paris-Saclay.
\bibliography{reference.bib}

\section*{Checklist}



 \begin{enumerate}

 \item For all models and algorithms presented, check if you include:
 \begin{enumerate}
   \item A clear description of the mathematical setting, assumptions, algorithm, and/or model. [Yes]
   \item An analysis of the properties and complexity (time, space, sample size) of any algorithm. [Yes]
   \item (Optional) Anonymized source code, with specification of all dependencies, including external libraries. [Yes]
 \end{enumerate}

 \item For any theoretical claim, check if you include:
 \begin{enumerate}
   \item Statements of the full set of assumptions of all theoretical results. [Yes]
   \item Complete proofs of all theoretical results. [Yes]
   \item Clear explanations of any assumptions. [Yes]     
 \end{enumerate}

 \item For all figures and tables that present empirical results, check if you include:
 \begin{enumerate}
   \item The code, data, and instructions needed to reproduce the main experimental results (either in the supplemental material or as a URL). [Yes]
   \item All the training details (e.g., data splits, hyperparameters, how they were chosen). [Yes]
         \item A clear definition of the specific measure or statistics and error bars (e.g., with respect to the random seed after running experiments multiple times). [Yes]
         \item A description of the computing infrastructure used. (e.g., type of GPUs, internal cluster, or cloud provider). [Yes]
 \end{enumerate}

 \item If you are using existing assets (e.g., code, data, models) or curating/releasing new assets, check if you include:
 \begin{enumerate}
   \item Citations of the creator If your work uses existing assets. [Yes]
   \item The license information of the assets, if applicable. [Not Applicable]
   \item New assets either in the supplemental material or as a URL, if applicable. [Not Applicable]
   \item Information about consent from data providers/curators. [Not Applicable]
   \item Discussion of sensible content if applicable, e.g., personally identifiable information or offensive content. [Not Applicable]
 \end{enumerate}

 \item If you used crowdsourcing or conducted research with human subjects, check if you include:
 \begin{enumerate}
   \item The full text of instructions given to participants and screenshots. [Not Applicable]
   \item Descriptions of potential participant risks, with links to Institutional Review Board (IRB) approvals if applicable. [Not Applicable]
   \item The estimated hourly wage paid to participants and the total amount spent on participant compensation. [Not Applicable]
 \end{enumerate}

 \end{enumerate}

\includepdf[pages={1-7}]{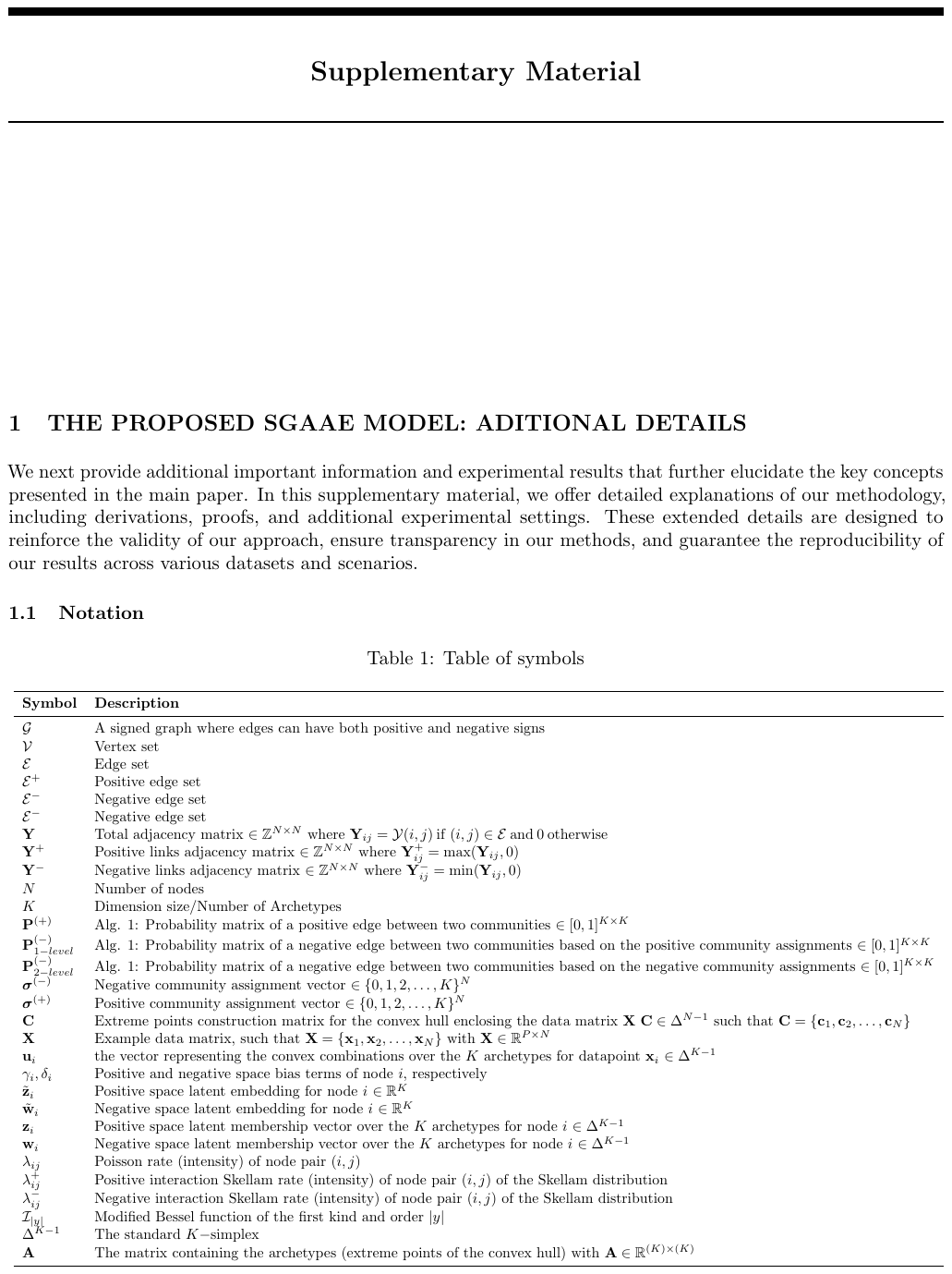}

\end{document}

%% file: 1-introduction.tex
\section{Introduction}
Graphs are commonly used to model complex relations and interactions between objects.
Thus, different types of real-world data, such as molecules and social networks, can be naturally modeled as graphs.
In many applications, we need to apply machine learning techniques on graphs.
For instance, predicting various properties of molecules (\eg quantum mechanical properties)~\citep{gilmer2017neural} has become an emerging topic in chemoinformatics.
This need for machine learning methods that operate on graphs led to the development of the field of graph representation learning~\citep{hamilton2020graph}.
In the past few years, interest in this field has flourished. 
Graph representation learning is mainly comprised of Graph Neural Networks (GNNs)~\citep{wu2020comprehensive}.
These models have become the standard tool for performing machine learning tasks on graphs.
Roughly speaking, GNNs learn vector representations of nodes (and potentially of graphs) in a supervised, end-to-end fashion.
So far, GNNs have been mainly evaluated in supervised learning tasks while
unsupervised learning of node representations, however, has not received the same amount of attention.

Existing methods for unsupervised learning of node representations typically employ an autoencoder (AE) framework.
In AEs, the encoder corresponds to a GNN that aggregates the local information of nodes, while the decoder reconstructs the entire graph from the learned node representations.
Most models are actually Variational Autoencoders (VAEs) which utilize latent variables.
The latent variables are usually formulated as Gaussian variables, while some loss term is employed to encourage them to be similar to some predefined prior distribution, typically an isotropic Gaussian distribution with diagonal covariance matrix.
To make models more interpretable, some works have replaced the Gaussian distribution with others, \eg Dirichlet distributions such that the latent variables describe graph cluster memberships~\citep{li2020dirichlet}. 
Due to their success for standard graphs, graph AEs and graph VAEs have been generalized to other types of graphs such as directed graphs~\citep{salha2019gravity,kollias2022directed} and hypergraphs~\citep{fan2021heterogeneous}.
However, prior work has mainly focused on unsigned graphs (where all edges are positive edges) even though signed graphs (\ie graphs in which each edge has a positive or negative sign) are ubiquitous in the real world~\citep{leskovec2010signed}.

In this paper, we propose a new model, the so-called Signed Graph Archetypal Autoencoder (SGAAE), which can embed nodes of signed graphs into vectors.
Those representations capture the polarization that occurs within the graph.
The different poles correspond to the corners of a polytope.
To encode the polarization dynamics, the model employs a likelihood function for signed edges based on the Skellam distribution~\citep{skellam1946frequency}, \ie the discrete probability distribution of the difference between two independent Poisson random variables. 

The paper also introduces the concept of \textsc{2-level} polarization. Typically, polarization is understood as a simple ``for or against" dynamic, which we characterize in terms of our work as a traditional \textsc{1-level} view. In this perspective, people are grouped into two opposing sides (e.g., left vs. right in a political debate), and polarization is defined by strong agreement within each group. For instance, members of team A generally like and agree with each other while disagree with members of team B. However, reality is much more nuanced. Imagine characterizing dynamics in a political debate: while we can easily identify left and right clusters, focusing only on this first level of polarization misses an important layer of complexity. The second level of polarization highlights how negative connections or disagreements, and animosities form their own distinct structures. This is not just about team A disapproving team B. Within team A, for example, there could be subgroups that are characterized with mutual animosity over something else entirely, even if they broadly agree on the main issue. In other words, a hidden network within the network emerges, fueled by negativity. This second level reveals a deeper community structure, formed not by agreement, but by shared animosity. These hidden layers of negativity play a critical role in shaping the overall dynamics of polarization.

The contributions of this work are summarized as follows:
\begin{itemize}
    \item \textit{A novel AE for signed networks.} We introduce a carefully designed signed graph AE that exploits polarization to encode the underlying interactions between nodes in the input graphs. The choice of the latent space dynamics provides an identifiable, interpretable and natural representation of node memberships, which is suitable for social network settings, and especially for addressing multiple levels of polarization. 
    \item \textit{Superiority in real-world downstream tasks and visualization tasks.}  We experimentally demonstrate that \textsc{SGAAE} significantly outperforms several baselines in the task of signed link prediction on real-world networks, while the learned polytope space allows for successful archetypal extraction and characterization.
    
    \item \textit{Introduction of the \textsc{2-level} network polarization.} Real signed networks contain polarized groups formed independently through positive or negative link structures, yielding a \textsc{2-level} polarization problem. Existing models typically focus on polarization scenarios where groups are constrained to be formed uniquely based on very dense intra-group positive ties and at the same time dense negative inter-group ties (we refer to this as \textsc{1-level} polarization). In the proposed \textsc{2-level} polarization scenario, this constraint is lifted, allowing polarization to emerge separately from the structures defined by positive and negative links.

    \item \textit{We show that the porposed AE can capture the \textsc{2-level} network polarization.} We demonstrate that our proposed framework enables a node to possess two sets of embedding vectors describing positive and negative group memberships,
effectively capturing both \textsc{1-level} and \textsc{2-level} polarization settings.
    
\end{itemize}


%% file: 2-related_work.tex
\section{Related Work}
\label{sec:related} 

\paragraph{Signed graph representation learning.}
Initial attempts for learning node representations in signed networks were inspired by advancements in the field of natural language processing (NLP), through embeddings on some vector space in an unsupervised manner.
For example, \textsc{SNE} adapts the log-bilinear model from the field of NLP such that the learned node representations capture node's path and sign information~\citep{yuan2017sne}.
Other embedding methods were designed to maintain structural balance, based on the assumption that triangles with an odd number of positive edges are more plausible than those with an even number.
For instance, \textsc{SIGNet} builds upon the traditional word2vec family of embedding approaches~\citep{mikolov2013distributed}, but replaces the standard negative sampling approach with a new method which maintains structural balance in higher-order neighborhoods~\citep{islam2018signet}.
Likewise, SiNE employs a multi-layer neural network to learn the node representations by optimizing an objective function satisfying structural balance theory~\citep{wang2017signed}.
Based on ideas from balance theory, \textsc{SGCN} generalizes the prominent Graph Convolutional Network~\citep{kipf2017semi} to signed graphs~\citep{derr2018signed}.
\textsc{SNEA} replaces the mean-pooling strategy of SGCN with an attention mechanism which allows it to aggregate more important information from neighboring nodes based on balance theory~\citep{li2020learning}.
SIDE extends random walk-based embedding algorithms to signed graphs~\citep{kim2018side}.
Embeddings are learned by maximizing the likelihood over both direct and indirect signed links.
\textsc{POLE} puts more focus on negative links than other works~\citep{huang2022pole} and uses signed autocovariance to capture topological and signed similarities.
To address some limitations of structural balance theory (\eg it cannot handle directed graphs), \textsc{SSNE} leverages the status theory and uses different translation strategies in the embedding space for positive and negative links~\citep{lu2019ssne}.
\textsc{SiGAT} is a motif-based variant of the Graph Attention Network where the extracted motifs model the two aforementioned classic theories in sociology, \ie balance theory and status theory~\citep{huang2019signed}. Finally, several recent studies in signed graph representation learning have integrated the signed graph Laplacian with GNNs, leading to the development of signed-spectral GNNs \citep{lap_1,lap_2_baseline,lap_3,lap4}.


\paragraph{Graph autoencoders.}
Owing to their simple structure, graph autoencoders (AEs) have been widely used in many domains to learn representations of nodes and/or graphs in an unsupervised manner.
Here, we focus on node-level representations.
We should mention though that there is a line of work that focuses on learning embeddings of graphs~\citep{winter2021permutation} or specific classes of graphs~\citep{zhang2019d}, mostly applied to molecular generation~\citep{simonovsky2018graphvae,jin2018junction}. 
Graph variational autoencoders (VAEs), instead of embedding each node into a vector as standard AEs, embed each node into a distribution.
The prior distribution over the latent features of nodes needs to be chosen a priori.
Most models employ the standard isotropic Gaussian distribution as their prior~\citep{kipf2016variational,grover2019graphite} or a Gaussian mixture
distribution~\citep{yang2019deep}.
Other models utilize more complex distributions such as Dirichlet distributions~\citep{li2020dirichlet} or the Gamma distribution~\citep{sarkar2020graph}.
The latent variables that emerge in those cases are more interpretable.
In the case of the Dirichlet distributions, they correspond to graph cluster memberships, while Gamma-distributed latent variables result in non-negativity
and sparsity of the learned embeddings and can also be considered as community memberships.
Typically, graph AEs and VAEs employ some GNN as their encoder~\citep{kipf2016variational,grover2019graphite}.
These can be modified to better capture specific properties of graphs, such as the community structure~\citep{salha2022modularity}.

The closest work to the proposed \textsc{SGAAE} is the Signed relational Latent dIstance Model (\textsc{SLIM}) \citep{SLIM} which is a latent distance-based model that extracts a unified embedding space based on the Skellam likelihood. Differently from our method, \textsc{SLIM} does not consider neural network-based representations, and it is not able to characterize \textsc{2-level} polarization scenarios. Moreover, our work combines graph neural networks with classical statistical models. To the best of our knowledge our paper acts as the first signed graph autoencoder that defines a self-explainable latent space.


%% file: 3-methods.tex
\section{Proposed Method}
\label{sec:method}

\label{methods}
\begin{figure*}
    \centering
    \includegraphics[width=0.8\textwidth]{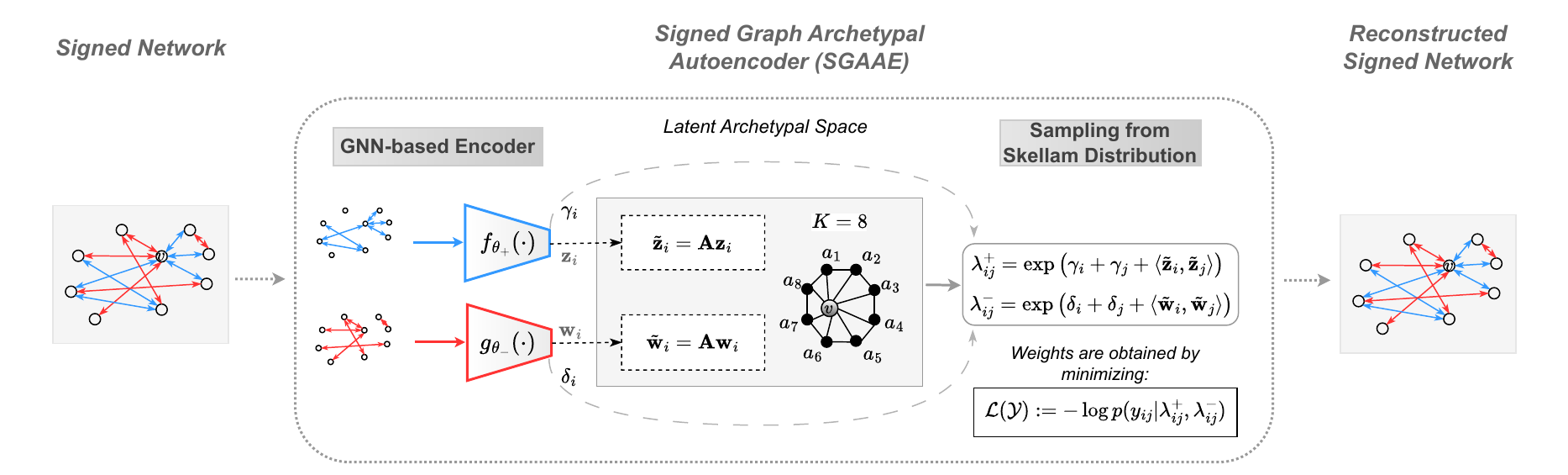}
    \caption{Framework of the proposed Signed Graph Archetypal Autoencoder (SGAAE). Given a signed network as input, the model utilizes two GNN-based encoding components working on the positive and negative interactions, respectively. The archetypal membership matrices $\bm{z}_i,\bm{w}_i$ are procured which are later multiplied with the archetypal matrix $\bm{A}$ to get the final node embeddings for the positive and negative spaces $\Tilde{\bm{z}}_i,\Tilde{\bm{w}}_i$. Then the final embeddings are used to calculate the Skellam rates optimizing for the Skellam log-likelihood for reconstructing the original signed graph.}
    \label{fig:sgae}
\end{figure*}
\paragraph{Preliminaries.}
Let \(\mathcal{G}=(\mathcal{V}, \mathcal{Y})\) be a \textit{signed graph}, where \(\mathcal{V}=\{1,\ldots,N\}\) represents the set of nodes, and \(\mathcal{Y}:\mathcal{V}^2 \rightarrow \mathsf{X} \subseteq \mathbb{R}\) is the map that assigns weights to pairs of nodes. An edge \((i,j)\in\mathcal{V}^2\) exists if the weight \(\mathcal{Y}(i,j)\) is non-zero. Thus, the set of edges in the network is given by \(\mathcal{E} := \{(i,j) \in \mathcal{V}^2 : \mathcal{Y}(i,j) \neq 0\}\). 
Considering the general case, we assume that a node pair relationship is characterized by integer values, \ie \(\mathsf{X} \subset \mathbb{Z}\). 
For simplicity, we focus on undirected graphs yielding \(\mathcal{Y}(i,j)\)=\(\mathcal{Y}(j,i)\), but we note here that our method easily generalizes to directed networks as well. We denote by $\mathcal{E}^{+}$ the positive edge set \(\{(i,j) \in \mathcal{V}^2 : \mathcal{Y}(i,j) > 0\}\) and by $\mathcal{E}^{-}$ the negative edge set \(\{(i,j) \in \mathcal{V}^2 : \mathcal{Y}(i,j) < 0\}\). Finally, we introduce the adjacency matrices $\bm{Y}$, $\bm{Y}^+$, and $\bm{Y}^- \in \mathbb{Z}^{N \times N}$ with $\bm{Y}_{ij}=\mathcal{Y}(i,j) \: \text{if } (i,j) \in \mathcal{E} \: \text{and} \: 0 \: \text{otherwise}$, while $\bm{Y}_{ij}^+=\max(\bm{Y}_{ij},0)$ and $\bm{Y}_{ij}^-=\min(\bm{Y}_{ij},0)$.

We next present the Signed Graph Archetypal Autoencoder (\textsc{SGAAE}), which is depicted in Figure \ref{fig:sgae}. Our primary aim is to design a graph AE capable of learning two sets of latent embeddings that are sufficient for explaining the relationships, as well as, for characterizing the structure defined by both the positive and negative graph interactions, in a disentangled manner. Specifically, for a given signed network $\mathcal{G}=(\mathcal{V}, \mathcal{Y})$, we aim to learn two sets of low-dimensional vectors $\{\Tilde{\bm{z}}_{i}\}_{i\in\mathcal{V}}\in\mathbb{R}^{K}$, $\{\Tilde{\bm{w}}_{i}\}_{i\in\mathcal{V}}\in\mathbb{R}^{K}$ ($K \ll |\mathcal{V}|$), where each is responsible for explaining the link structure as observed by $\mathcal{E}^{+}$ and $\mathcal{E}^{-}$, respectively. Importantly, the model should take into account both representations when predicting the existence of a link along its sign, or when characterizing the combined structure of the signed graph. Furthermore, such projections should express node memberships over distinct extreme profiles existing in the network, referred to as archetypes, facilitating the understanding of network polarization in a straightforward and interpretable way. 

\begin{figure}[!t]
\centering
\begin{subfigure}{0.48\textwidth}  
    \centering
    \includegraphics[width=1\linewidth]{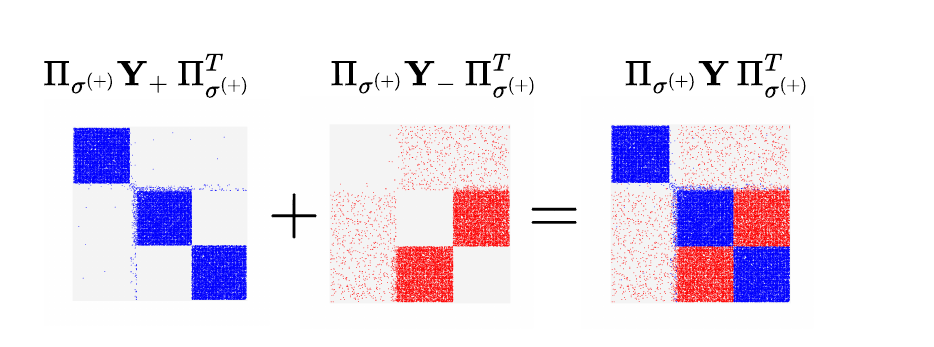}
    \caption{\textsc{1-level polarization}}
    \label{fig:1_pol}
\end{subfigure}
\hfill
\begin{subfigure}{0.48\textwidth}  
    \centering
    \includegraphics[width=1\linewidth]{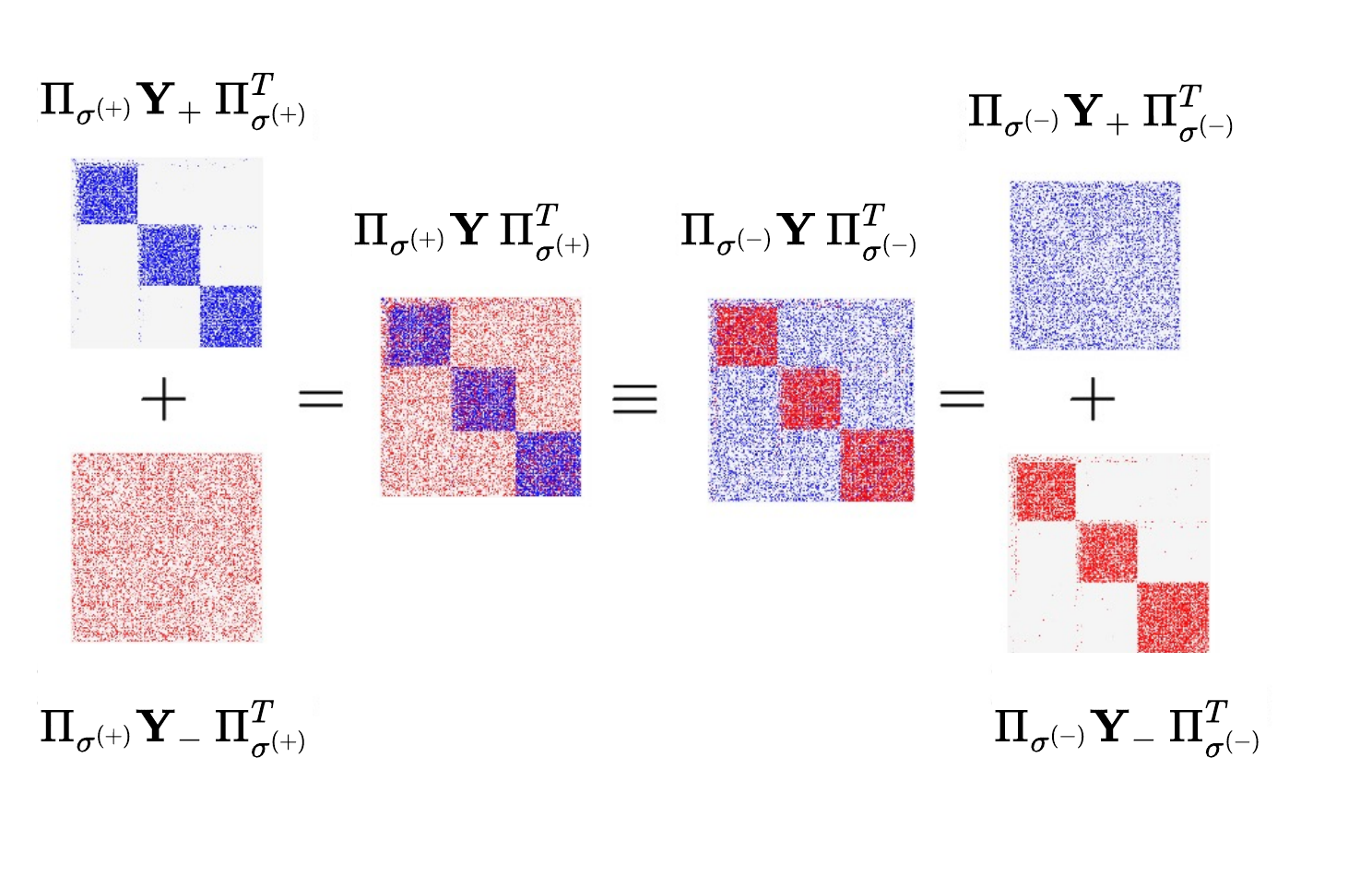}
    \caption{\textsc{2-level polarization}}
    \label{fig:2_pol}
\end{subfigure}
\caption{\textsc{Different levels of network polarization:} Blue elements define positive ties whereas red define negative ties, displaying the total signed networks broken down into its two sign-specific components. \textbf{(a)} Showcases the traditional \textsc{1-level} definition of network polarization focused on extracting structures with dense intra-community positive connections and negative inter-community ties. The adjacency matrices $\bm{Y}$, $\bm{Y}^+$, and $\bm{Y}^-$ and here are re-ordered based on the permutation matrix $\bm{\Pi}_{\bm{\sigma}^{(+)}}$ under the positive community memberships $\bm{\sigma}^{(+)}$. \textbf{(b)} Showcases two different permutations of the same signed network and its corresponding components that can be broken down under very dense positive communities and very dense negative communities. Importantly, the community memberships in the network for the negative and positive structures are not the same, yielding a \textsc{2-level} polarization. The adjacency matrices in the left panel are re-ordered based on the permutation matrix $\bm{\Pi}_{\bm{\sigma}^{(+)}}$ under the positive community memberships $\bm{\sigma}^{(+)}$ while in the right panel we re-order $\bm{Y}$, $\bm{Y}^+$, and $\bm{Y}^-$ based on the permutation matrix $\bm{\Pi}_{\bm{\sigma}^{(-)}}$ under the negative community memberships $\bm{\sigma}^{(-)}$.}
\label{fig:2pols}
\end{figure}

\textbf{The \textsc{2-level} Network Polarization.} 
Quantifying network polarization has been long tied to antagonistic group mining in literature \citep{pol1,pol2,pol3,pol4,pol5}. Specifically, there has been great focus in community detection to discover groups defining \textbf{(i)} \textit{strong intra-community ties}, \ie nodes within the same community that have dense, positive connections with each other, reflecting high levels of agreement, similarity, or cooperation, and \textbf{(ii)}\textit{ weak or negative inter-community ties}, \ie connections between nodes from different communities that are sparse, weak, or negative, indicating disagreement, conflict, or lack of interaction. We define this setting as \textsc{1-level} polarization. This is illustrated in Figure~\ref{fig:1_pol} where the positive structure (blue points) is well separated from the negative structure (red points) that creates the \textsc{1-level} polarized network. Such a case disregards potential structure in the negative ties, \ie hidden disagreement/animosity structures, despite a uniform appearing inter-level negative tie pattern. An example of this is shown in Figure~\ref{fig:2_pol} where the same single network is re-ordered based on the community memberships that emerge from both the positive link structure (as in \textsc{1-level} case), but also from the negative link structure. We argue that looking solely at the positive structure makes a model completely blind to the negative, highly polarized structure. We define this case as an example of \textsc{2-level} polarization. A generative process of the multi-level polarized network is provided in Algorithm~\ref{alg:2-level}. 
Specifically, for a signed network with $N$ number of nodes, we define a positive community assignment vector $\bm{\sigma}^{(+)}\in \{0,1,2,\dots,K\}^{N}$, and a negative community assignment vector $\bm{\sigma}^{(-)} \in \{0,1,2,\dots,K\} 
^N$.
In addition, three probability matrices are given as input; \textbf{(i)} a positive community probability matrix $\bm{P}^{(+)} \in [0,1]^{K\times K}$ that denotes the probability of a positive edge between two communities, \textbf{(ii)} the \textsc{1-level} negative community probability matrix $\bm{P}^{(-)}_{1-level}\in [0,1]^{K\times K}$ that denotes the probability of a negative edge between two communities based on the positive community assignments, and \textbf{(iii)}  the \textsc{2-level} negative community probability matrix $\bm{P}^{(-)}_{2-level}\in [0,1]^{K\times K}$ that denotes the probability of a negative edge between two communities based on the negative community assignments. 
Finally, we introduce the polarization probability scalar $0\leq\alpha\leq 1$ which controls the polarization level, \ie how a negative edge is generated either based on the positive community vector $\bm{\sigma}^{(+)}$ (higher $\alpha$ values), or either via the independent negative community structure $\bm{\sigma}^{(-)}$ (lower $\alpha$ values). 
In Figure~\ref{fig:1_pol} we set $\alpha=0.95$, and in Figure~\ref{fig:2_pol} we set $\alpha=0.05$. (More details in Subsection 1.6 in the supplementary.)

\begin{algorithm}[!t]
\scriptsize
\caption{Signed Network Generation for \textsc{1-level} and \textsc{2-level} Polarized Networks}
\label{alg:2-level}
\begin{algorithmic}[1]
\State \textbf{Input:}
 $N$, $\bm{\sigma}^{(+)}$, $\bm{\sigma}^{(-)}$, $\mathbf{P}^{(+)}$,$\mathbf{P}^{(-)}_{1-level}$, $\mathbf{P}^{(-)}_{2-level}$, $\alpha$.

\State Initialize two $N \times N$ zero matrices $\mathbf{A}^{(+)}$ and $\mathbf{A}^{(-)}$ that represent graphs $G^{(+)}$ and $G^{(-)}$.
\For{$i = 1$ to $N$}
    \For{$j = i+1$ to $N$}
        \If {$rand() < \mathbf{P}^{(+)}[\sigma^{(+)}_i, \sigma^{(+)}_j]$}
            \State $\mathbf{A}_{ij}^{(+)} \gets 1$
            \State $\mathbf{A}_{ji}^{(+)} \gets 1$
                
        \Else

        \If {$rand() < \alpha$} 
      
            \If {$rand() < \mathbf{P}^{(-)}_{1-level}[\sigma^{(+)}_i, \sigma^{(+)}_j]$}
                 \State $\mathbf{A}_{ij}^{(-)} \gets 1$
                 \State $\mathbf{A}_{ji}^{(-)} \gets 1$
            \EndIf
        \Else
    
            \If {$rand() < \mathbf{P}^{(-)}_{2-level}[\sigma^{(-)}_i, \sigma^{(-)}_j]$}
                \State $\mathbf{A}_{ij}^{(-)} \gets 1$
                \State $\mathbf{A}_{ji}^{(-)} \gets 1$
                
            \EndIf
        \EndIf
    \EndIf

    \EndFor
\EndFor
\State $\mathbf{A} \gets \mathbf{A}^{(+)} - \mathbf{A}^{(-)}$
\State \Return $\mathbf{A}$ \Comment{adjacency matrix of generated graph}  
\end{algorithmic}
\end{algorithm}

Based on the above, we next combine and generalize recent advances in signed graph representation learning \citep{SLIM} with GNNs, to design a graph AE based on an explainable latent space that confirms the effectiveness of network multi-level polarization targeting the analysis of social networks.

\textbf{Archetypal Analysis.} The extreme points of the convex hull enclosing the data can be referred to as archetypes. 
Archetypes, essentially represent the vertices of the convex hull defined by the data and can thus provide a detailed view of its inherent extreme structural traits.
Therefore, archetypal analysis (AA)~\citep{AA1,AA2} can allow us to identify and explain core data dependencies while extracting data representations as convex combinations of extremal points.

Let $\bm{X}\in\mathbb{R}^{P\times N}$ a data matrix, such that $\bm{X}=\{\bm{x}_1,\bm{x}_2,\dots,\bm{x}_N\}$.
Then, the archetype matrix $\bm{A} \in\mathbb{R}^{P\times K}$, $\bm{A}=\{\bm{a}_1,\bm{a}_2,\dots,\bm{a}_K\}$, where $K\ll P$, can be extracted as follows:
\begin{equation}
    \bm{\alpha}_j=\sum_{i=1}^N \bm{x}_i c_{ij},
\end{equation}
\noindent where $\boldsymbol{c}_j\in \Delta^{N-1}$ with $\Delta^{N-1}$ denoting the $N$-dimensional standard simplex, such as $c_{ij}\geq 0$ and $\sum_i c_{ij}=1$. 
Let $\bm{A}$ be the convex hull of the data, then each point $\bm{x}_i$ is reconstructed as follows:
\begin{equation}
    \bm{x}_i=\sum_{j=1}^{K} \bm{a}_j u_{ji}
\end{equation}
\noindent where $\boldsymbol{u}_i^T\in \Delta^{K-1}$ denoting the standard simplex in $K$-dimensions. Let $\bm{U} \in\mathbb{R}^{N\times K}$ the matrix formed by the $K$-dimensional embeddings for each data point. Then, $\bm{U}$ captures the representations as convex combinations of the archetypes in $\bm{A}$, as follows:
\begin{eqnarray}
    \bm{X}\approx \bm{XCU}\quad
    \text{s.t. }\boldsymbol{c}_j\in \Delta^{N-1} \text{ and } \bm{u}_i^T\in \Delta^{K-1}.
\end{eqnarray}
In the above, archetypes are represented by the corners of the convex hull, denoted as $\bm{A}=\bm{X}\bm{C}$.

\textbf{The Likelihood.} Following \cite{SLIM}, we use the Skellam distribution, which models the difference of two independent Poisson-distributed random variables ($y=N_1 - N_2\in\mathbb{Z}$) with rates $\lambda^{+}$ and $\lambda^{-}$ as: 
\begin{align*}
P(y|\lambda^{+},\lambda^{-}) = e^{-(\lambda^{+}+\lambda^{-})}\left(\frac{\lambda^{+}}{\lambda^{-}}\right)^{y/2}\mathcal{I}_{|y|}\left(2\sqrt{\lambda^{+}\lambda^{-}}\right),
\end{align*}
where $N_1 \sim Pois(\lambda^{+})$ and $N_2 \sim Pois(\lambda^{-})$, and $\mathcal{I}_{|y|}$ is the modified Bessel function of the first kind and order $|y|$. In particular, $\lambda^{+}$ generates the intensity of positive outcomes for $y$ while $\lambda^{-}$ generates the intensity of negative outcomes. Thus, the negative log-likelihood, used as our loss function, is computed as:
\begin{align}
\mathcal{L}(\mathcal{Y}) &:=-\log p(y_{ij}|\lambda^{+}_{ij},\lambda^{-}_{ij}) \\
&= \sum_{i<j}{(\lambda^{+}_{ij}+\lambda^{-}_{ij})} - \frac{y_{ij}}{2}\log\left(\frac{\lambda^{+}_{ij}}{\lambda^{-}_{ij}}\right)-\log(I_{ij}^{*}),\label{eq:loss_fun}
\end{align}
where $I_{ij}^{*} := \mathcal{I}_{|y_{ij}|}\left(2\sqrt{\lambda^{+}_{ij}\lambda^{-}_{ij}}\right)$. 

For relational data, the Skellam distribution rate parameter $\lambda^{+}_{ij}$ models the positive interaction intensity and $\lambda^{-}_{ij}$ parameter models negative the interaction intensity between a node pair $\{i,j\}$. We constrain the latent space into a polytope to define a convex hull of the latent representations, achieving archetypal characterization. For relational data, we aim on expressing the embedding matrices—which encode each node's position in the network—as convex combinations of nodal archetypes. Consequently, we follow a similar methodology as the \textsc{SLIM} model, extending the relational archetypal analysis formulation to two membership vectors in the same latent spaces, independently reconstructing positive and negative representations from the extracted archetypes. This enables us to disentangle memberships in positive and negative communities, allowing for the characterization of \textsc{2-level} polarization. Thus, the Skellam rates are calculated as:
\begin{align}
    \lambda_{ij}^{+} &=\exp \big( \gamma_{i} + \gamma_{j} +\langle\Tilde{\bm{z}}_{i},\Tilde{\bm{z}}_{j}\rangle\big),\label{inner1}
    \\
\lambda_{ij}^{-} &=\exp \big( \delta_{i} + \delta_{j} +\langle\Tilde{\bm{w}}_{i},\Tilde{\bm{w}}_{j}\rangle\big),\label{inner2}
\end{align}
where $\Tilde{\bm{z}}_i,\Tilde{\bm{w}}_i \in\mathbb{R}^{K}$ the node embedding positions, $\langle \cdot,\cdot \rangle$ denotes the inner product while $\{\gamma_i,\delta_i\}_{i\in\mathcal{V}}$ denote the node-specific random effect terms. Essentially, $\gamma_i,\gamma_j$ represent the tendency of a node to form positive connections while
$\delta_i,\delta_j$ the tendency to form negative connections. In other words, it accounts for degree heterogeneity in the positive and negative sub-networks, accordingly. To further construct explainable latent spaces, defined via memberships over archetypes, the final embedding positions $\Tilde{\bm{z}}_i,\Tilde{\bm{w}}_i$ are obtained as follows:
\begin{align}
    \Tilde{\bm{z}}_{i} &=\bm{A}\bm{z}_{i},
    \\
\Tilde{\bm{w}}_{i} &=\bm{A}\bm{w}_{i},\label{LRPM_inensity_function_2}
\end{align}
where $\bm{z}_i,\bm{w}_i\in \Delta^{K-1}$ and $\bm{A} \in\mathbb{R}^{K\times K}$ is the matrix with columns containing the archetypes for the unified space of positive and negative representations, that are reconstructed through the membership vectors. To define memberships that utilize archetypes to the maximum, \ie archetypes belonging to the data, we introduce a temperature parameter in the softmax function used to define $\bm{z}_i,\bm{w}_i$. There exist additional methods that define the archetypal matrix through a gate function to achieve the same effect~\citep{SLIM}. (A comprehensive list of all introduced notation is provided in Table 1 of the supplementary material.)

The latent space is self-explainable by construction, as each node's position in the latent space describes its membership to the $K$ introduced archetypes. This approach ensures interpretability, allowing us to understand and explain the role and position of each node within the network. Our methods build on the principles of explainability demonstrated in prior works~\citep{expl1,expl2,expl3}, where self-explainability is achieved by introducing prototypes in the latent space, a concept closely related to the relational archetypal analysis adopted in our model.

\begin{figure}[b]
    \centering
    \includegraphics[width=.4\textwidth]{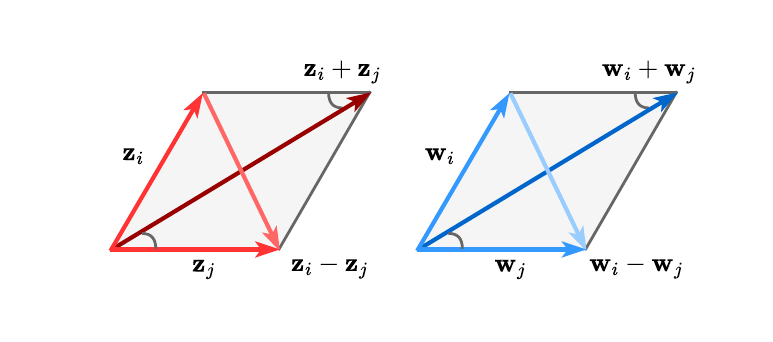}
    \captionof{figure}{Polarization Identity---Inner product proximity in Skellam rates.}
    \label{fig:inner-product-skellam}
\end{figure}
\textbf{Characterizing \textsc{2-level} Polarization.} In real signed networks, we may have communities that are created based on positive links (\textit{friendship}) and additional communities that are created due to negative links (\textit{animosity}). Models that exploit \textsc{1-level} polarization essentially introduce only one vector of community memberships, which constrains the node to belong to one community under both the positive and negative structures. In a \textsc{2-level} polarization scenario, this constraint is removed, by enabling a node to have two sets of embedding vectors, each one describing the community structure based on either positive or negative link structures, separately. This is evident in our framework since we introduce two archetype membership matrices $\mathbf{Z},\mathbf{W}$. In Figure \ref{fig:2_pol}, the same unique signed network is defined and characterized by \textsc{2-level} polarization and is re-ordered in two different ways. More specifically, the left part of Figure \ref{fig:2_pol} focuses on discovering structure based on finding densely positively connected communities with densely negative connections leaving the (exact same) communities. This essentially describes the \textsc{1-level} polarization scenario under a unique community membership vector $\mathbf{Z}$, and the adjacency matrices of the left panel are reordered based on exactly these community memberships (we show how both the positive and negative link structures are re-ordered as well as their combination on the same adjacency metric). As we witness, such a case finds very strong communities in the positive links but the negative structure seems spurious/random. The right panel of Figure \ref{fig:2_pol} shows exactly the same network while seeking communities based on the negative link structure (by introducing a community membership vector $\mathbf{W}$), \ie very negatively connected communities with positive links leaving the (exact same) communities. We observe that under this scenario, we successfully find very strong signals in the negative communities but the positive link structure seems random. 
 Therefore, there is a question of choice between the panels to effectively describe this very differently structured network, based on the positive/negative structure. 
 Motivated by this, our methodological design combines both possibilities by decoupling the community archetype memberships while accounting for this \textsc{2-level} polarization problem, as we also show in section \ref{sec:experiments}.

\textbf{GNN-based Encoder.} To handle the signed interactions in the network, we construct an encoder based on message-passing neural networks (MPNNs) for each edge type. Specifically, our approach separates the positive and negative interactions to learn distinct embeddings for each type. 
For the positive interactions, the encoder processes the graph containing only positive links by setting the weights of negative links to zero. For the negative interactions, we flip the sign of the negative links to treat them as positive during message passing.  
We use graph convolutional network (GCN) \citep{kipf2016semi} with each layer in this network propagating information among connected nodes, followed by a Multilayer Perceptron (MLP) to produce the final embeddings and parameters.
Thus, representations ${\bm{Z}, \bm{\gamma}}$ are produced for positive links, whereas ${\bm{W}, \bm{\delta}}$ are produced for the negative ones. (\textit{More details are provided in the supplementary in Subsection 2.3}.)

\textbf{Choosing the inner product.} We here adopt the inner product as a proximity metric for the Skellam rates (see Figure \ref{fig:inner-product-skellam}) instead of the Euclidean distance~\citep{SLIM}, as we argue that it can capture more complex relations since it weakly generalizes distance matrices (see supplementary). We can think of the archetype membership vectors $\bm{z}_i,\bm{w}_i$ as the participation to the extreme profiles/opinions present in the network of node $i$. For a real inner product space, by using the polarization identity we can express the inner product of Eq. \eqref{inner1} and \eqref{inner2} as:
\begin{equation}
\langle\Tilde{\bm{z}}_{i},\Tilde{\bm{z}}_{j}\rangle = \frac{1}{4} \left( \|\bm{A}(\bm{z}_i + \bm{z}_j)\|^2 - \|\bm{A}(\bm{z}_i - \bm{z}_j)\|^2 \right),
\end{equation}
\begin{equation}
\langle\Tilde{\bm{w}}_{i},\Tilde{\bm{w}}_{j}\rangle= \frac{1}{4} \left( \|\bm{A}(\bm{w}_i + \bm{w}_j)\|^2 - \|\bm{A}(\bm{w}_i - \bm{w}_j)\|^2 \right).
\end{equation}

The polarization identity can extend our understanding regarding the inner product as a similarity metric between two network nodes in terms of stochastic equivalence and homophily \citep{eigen}, two fundamental properties of social network analysis. For either vector $\mathbf{z}$, and $\mathbf{w}$, the first term of the polarization identity essentially describes that the similarity of two nodes is high when $ \|\bm{A}(\bm{z}_i + \bm{z}_j)\|^2$ is high, modeling stochastic equivalence. The second term denotes that a high rate can also be achieved when $ \|\bm{A}(\bm{z}_i - \bm{z}_j)\|^2$ is low meaning that the two nodes are positioned close to each other and thus modeling homophily. Optimizing the model using the polarization identity expression, yields a more stable optimization due to the nice properties and convexity of the squared Euclidean norm.

All of the introduced notation can be found in Table 1 of the supplementary material. Finally, our model scales efficiently to large networks by employing a random sampling technique; for further details, see subsection 1.4 in the supplementary material.


%% file: 4-experiments.tex
\section{Results and Discussion}
\label{sec:experiments}

We extensively evaluate the performance of our proposed method by comparing it to the prominent graph-based approaches designed for signed networks. All training details are provided in the supplementary.



\textbf{Datasets and baselines.} For the task of link prediction, the following \textit{real-world} benchmark datasets, shown in Table~\ref{tab:datasets}, are considered: (\textbf{1}) \textsl{Reddit} \citep{dataset_reddit}, (\textbf{2}) \textsl{Twitter} \citep{dataset_twitter}, (\textbf{3}) \textsl{wikiRfA} \citep{dataset_wikirfa}, and (\textbf{4}) \textsl{wikiElec} \citep{dataset_wikielec}. Furhtermore, we evaluate the performance of the proposed graph autoencoder in several tasks with comparisons against benchmark signed graph representation methods, including: 
(\textbf{i}) \textsc{POLE} \citep{huang2022pole}, (\textbf{ii}) \textsc{SLF} \citep{slf}, (\textbf{iii}) \textsc{SiGAT} \citep{huang2019signed}, (\textbf{iv}) \textsc{SIDE} \citep{kim2018side}, (\textbf{v}) \textsc{SigNet} \citep{islam2018signet}, (\textbf{v}) the Spectral-SGCN (\textsc{S-GCN}) \cite{lap_2_baseline}, and finally, (\textbf{vi}) \textsc{SLIM} \citep{SLIM}. (For Additional details see subsection 2.2 in the supplementary).



\begin{table*}[ht]
    \centering
    \begin{minipage}[t]{.34\linewidth}
        \vspace{0pt} 
        \centering
        \caption{Network statistics; \#Nodes, \#Positive, \#Negative links.}
        \label{tab:datasets}
        \resizebox{\linewidth}{!}{%
        \begin{tabular}{rcccc}
        \toprule
         & $|\mathcal{V}|$ & $|\mathcal{Y}^{+}|$ & $|\mathcal{Y}^{-}|$ & Density \\\midrule
        \textsl{Reddit} & 35,776 & 128,182 & 9,639 & 0.0001 \\
        \textsl{Twitter} & 10,885 & 238,612 & 12,794 & 0.0021 \\
        \textsl{wiki-Elec} & 7,117 & 81,277 & 21,909 & 0.0020 \\
        \textsl{wiki-RfA} & 11,332 & 117,982 & 66,839 & 0.0014\\\bottomrule
        \end{tabular}%
        }
    \end{minipage}%
    \hspace{0.02\linewidth}
    \begin{minipage}[t]{.6\linewidth}
        \vspace{0pt} 
        \centering
        \caption{Area Under Curve (AUC-PR) scores for representation size of $K=8$. (OOM: memory or high runtime error.) The standard
error of the mean for all the cases is approximately 0.005.}
        \label{tab:auc_pr}
        \resizebox{\linewidth}{!}{%
        \begin{tabular}{rccccccccccccccccccccccccc}\toprule
        \multicolumn{1}{l}{} & \multicolumn{3}{c}{\textsl{WikiElec}} & \multicolumn{3}{c}{\textsl{WikiRfa}} & \multicolumn{3}{c}{\textsl{Twitter}}& \multicolumn{3}{c}{\textsl{Reddit}} \\\cmidrule(rl){2-4}\cmidrule(rl){5-7}\cmidrule(rl){8-10}\cmidrule(rl){11-13}
        \multicolumn{1}{r}{Task} & $p@n$ & $p@z$ & $n@z$ & $p@n$ & $p@z$ & $n@z$ & $p@n$ & $p@z$ & $n@z$ & $p@n$ & $p@z$ & $n@z$  \\\cmidrule(rl){1-1}\cmidrule(rl){2-2}\cmidrule(rl){3-3}\cmidrule(rl){4-4}\cmidrule(rl){5-5}\cmidrule(rl){6-6}\cmidrule(rl){7-7}\cmidrule(rl){8-8}\cmidrule(rl){9-9}\cmidrule(rl){10-10}\cmidrule(rl){11-11}\cmidrule(rl){12-12}\cmidrule(rl){13-13}
        \textsc{POLE}    &.929 &.922 &.544 &.927 &.937 &.779 &\underline{.998} &.932 &.668 &OOM &OOM &OOM\\
        \textsc{SLF} &\textbf{.964} &.926 &\underline{.787} &\textbf{.983} &.922 &.881 &.994 &.870 &.740 &\textbf{.966} &.956 &.850\\
        \textsc{SiGAT}  &\underline{.960} &.724 &.439 &.969 &.646 &.497 &\textbf{.999} &.861 &.582 &\underline{.965} &.692 &.232\\
        \textsc{SIDE}    &.907 &.779 &.608 &.920 &.806 &.739 &.974 &.831 &.469 &.957 &.820 &.614\\
        \textsc{SigNet} &.944 &.670 &.298 &.950 &.572 &.417 &\underline{.998} &.647 &.248 &.956 &.510 &.083\\
        \textsc{S-SGCN} &.952 &.920 &.759 &.9630 &.911 &.853 &.989 &.917 &.601 &.960 &.935 &.806\\
        \textsc{SLIM}    &.953 &\underline{.956} &.785 &.973 &\underline{.969} &\underline{.907} &\textbf{.999} &\textbf{.962} &\underline{.813} &.958 &\textbf{.960} &.850\\
        \midrule
        \textsc{SGAAE}     &.960 &\textbf{.972} &\textbf{.895} &\underline{.980} &\textbf{.976} &\textbf{.951} &\textbf{.999} &\underline{.960} &\textbf{.816} &.964 &\underline{.961} &.\textbf{866}
        \\\bottomrule    
        \end{tabular}%
        }
    \end{minipage}
\end{table*}

\begin{figure*}[!t]
\centering
\begin{subfigure}{0.125\textwidth}  
    \centering
    \includegraphics[width=1\linewidth]{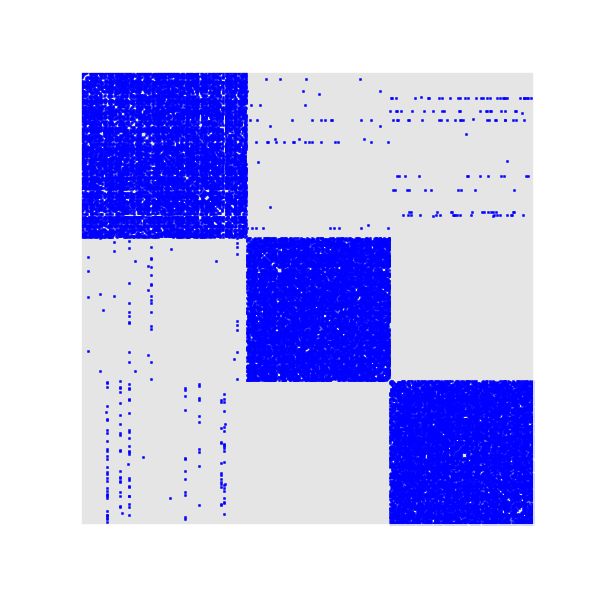}
    \caption{\tiny \textsc{SGAAE:1-level}}
    \label{fig:1_level_pol_1}
\end{subfigure}
\hfill
\begin{subfigure}{0.125\textwidth}  
    \centering
    \includegraphics[width=1\linewidth]{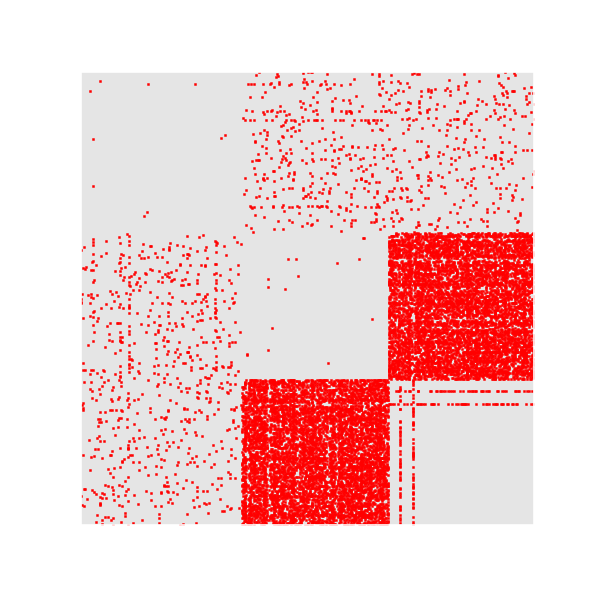}
    \caption{\tiny \textsc{SGAAE:1-level}}
    \label{fig:1_level_pol_2}
\end{subfigure}
\hfill
\begin{subfigure}{0.125\textwidth}  
    \centering
    \includegraphics[width=1\linewidth]{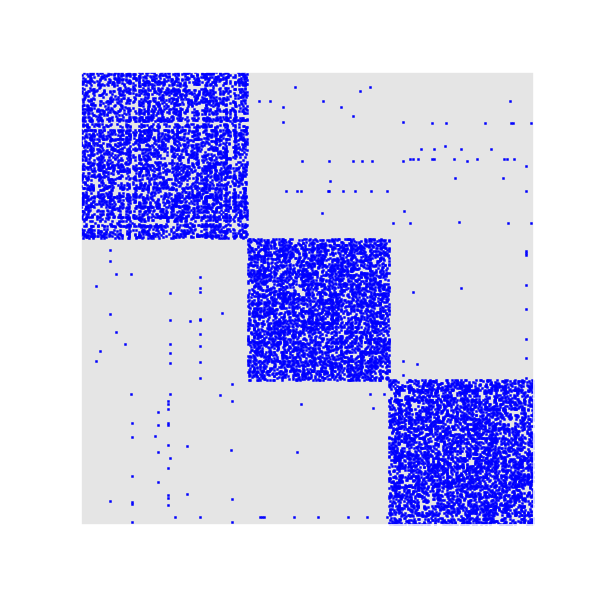}
    \caption{\tiny \textsc{SGAAE:2-level}}
    \label{fig:2_level_pol_1}
\end{subfigure}
\hfill
\begin{subfigure}{0.125\textwidth}  
    \centering
    \includegraphics[width=1\linewidth]{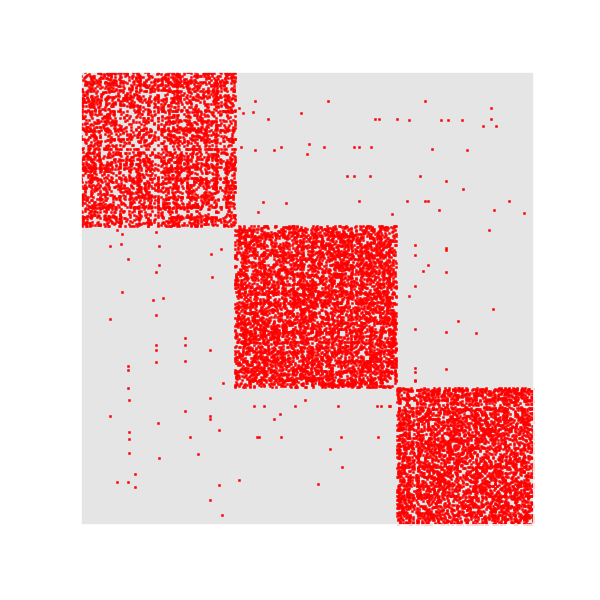}
    \caption{\tiny\textsc{SGAAE:2-level}}
    \label{fig:2_level_pol_2}
\end{subfigure}
\hfill
\centering
\begin{subfigure}{0.11\textwidth}  
    \centering
    \includegraphics[width=1.15\linewidth]{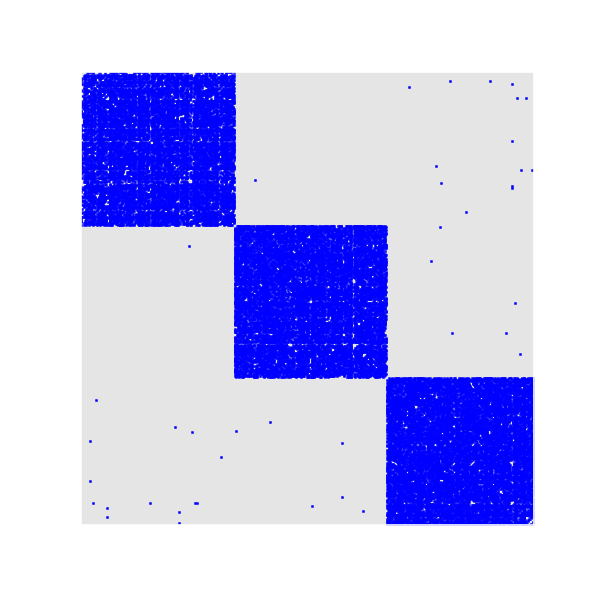}
    \caption{\tiny\textsc{SLIM:1-level}}
    \label{fig:1_level_pol_3}
\end{subfigure}
\hfill
\begin{subfigure}{0.11\textwidth}  
    \centering
    \includegraphics[width=1.15\linewidth]{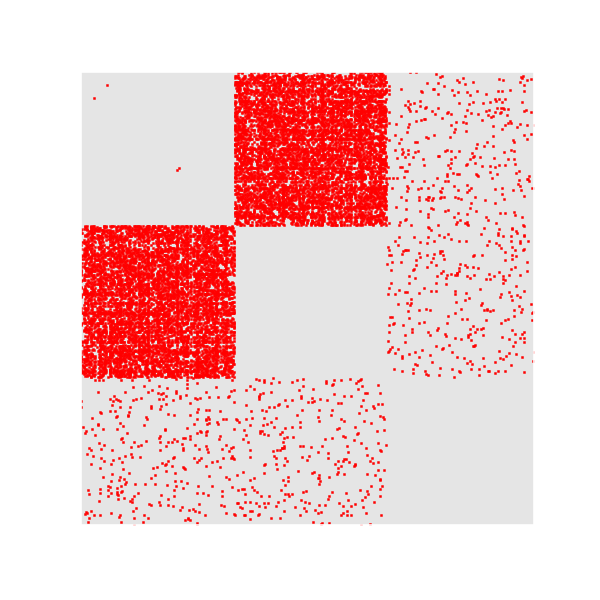}
    \caption{\tiny\textsc{SLIM:1-level}}
    \label{fig:1_level_pol_4}
\end{subfigure}
\hfill
\begin{subfigure}{0.11\textwidth}  
    \centering
    \includegraphics[width=1.15\linewidth]{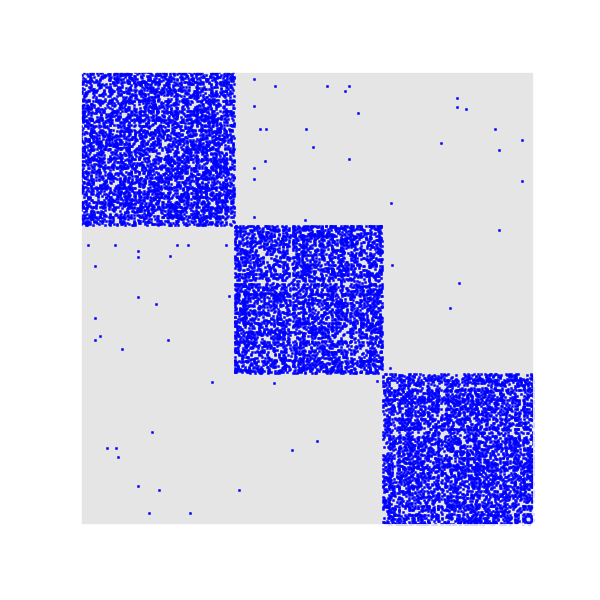}
    \caption{\tiny\textsc{SLIM:2-level}}
    \label{fig:2_level_pol_3}
\end{subfigure}
\hfill
\begin{subfigure}{0.11\textwidth}  
    \centering
    \includegraphics[width=1.15\linewidth]{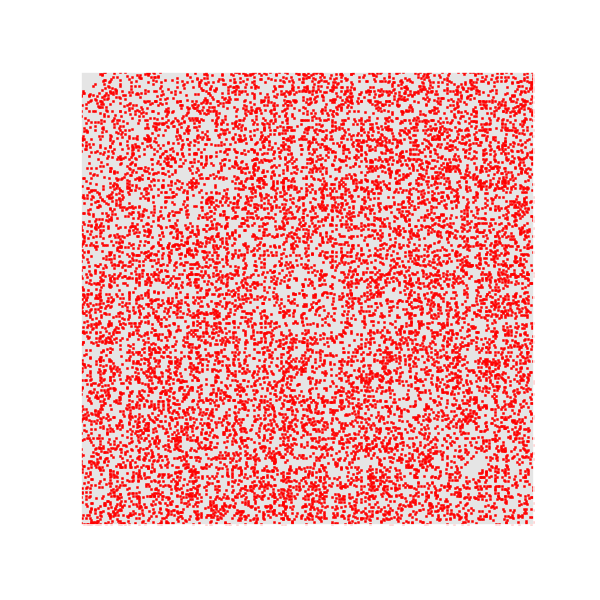}
    \caption{\tiny\textsc{SLIM:2-level}}
    \label{fig:2_level_pol_4}
\end{subfigure}
\caption{\textsc{Inferred community memberships:} (a)-(d) visualization of the re-ordered adjacency matrices based on the inferred community memberships of \textsc{SGAAE}, proving the expressive capabilities to provide characterization for both levels of polarization. (e)-(h) the same visualization for \textsc{SLIM} failing to explain or detect the structure over the negative ties and thus unable to account for the different polarization levels.}
\label{fig:models_2pols}
\end{figure*}

\begin{figure}[!h]
\centering
\begin{subfigure}[t]{0.23\textwidth}  
    \centering
    \includegraphics[width=0.9\linewidth]{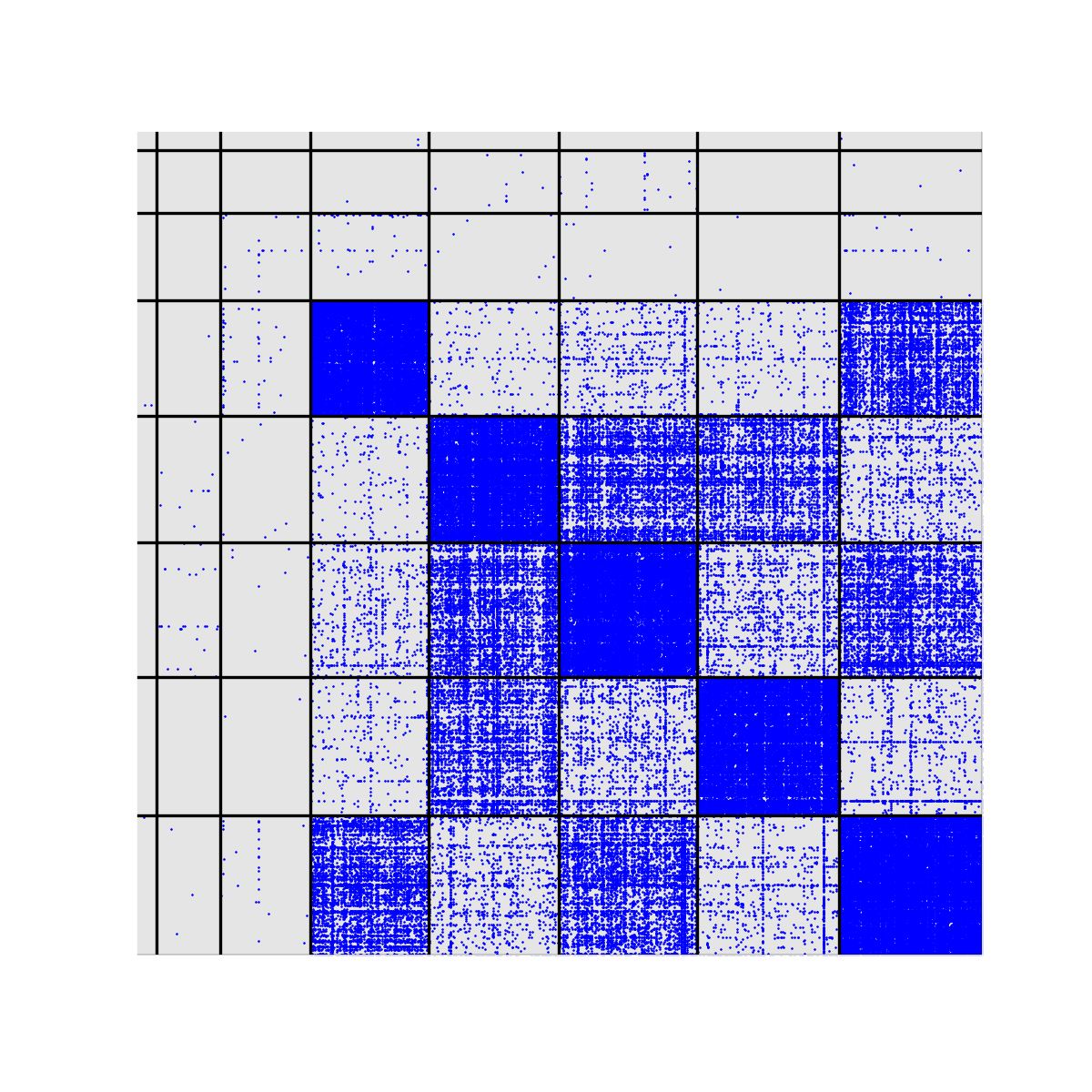}
    \caption{\scriptsize\textsc{Positive Adjacency}}
    \label{fig:dataset_pol_1}
\end{subfigure}
\hfill
\begin{subfigure}[t]{0.23\textwidth}  
    \centering
    \includegraphics[width=1\linewidth]{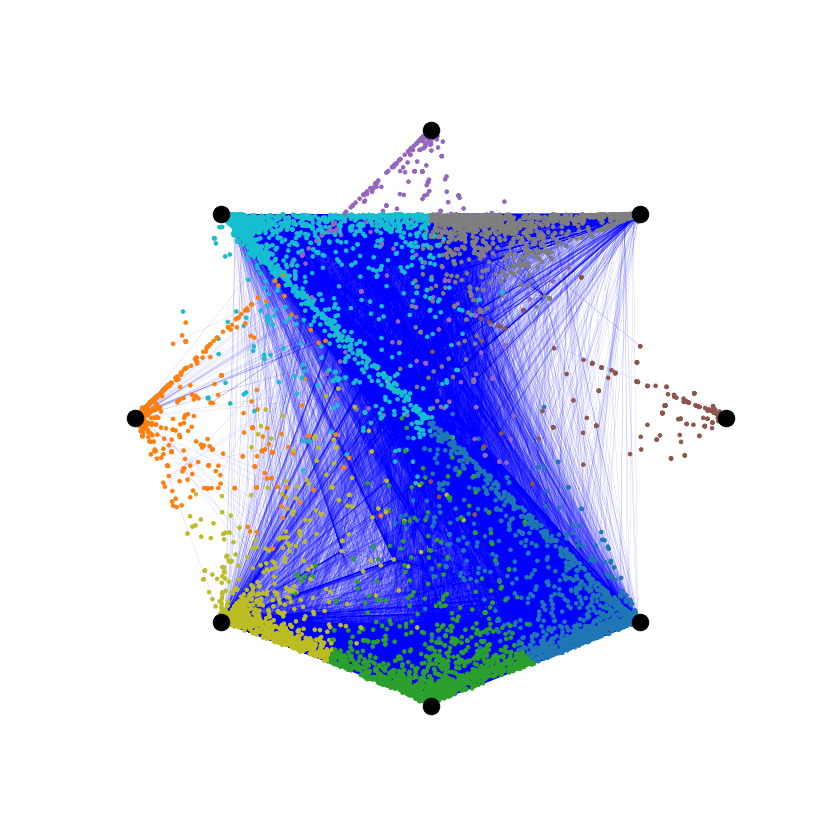}
    \caption{\scriptsize\textsc{Positive membership space}}
    \label{fig:dataset_pol_2}
\end{subfigure}
\hfill
\centering
\begin{subfigure}[t]{0.23\textwidth}  
    \centering
    \includegraphics[width=0.9\linewidth]{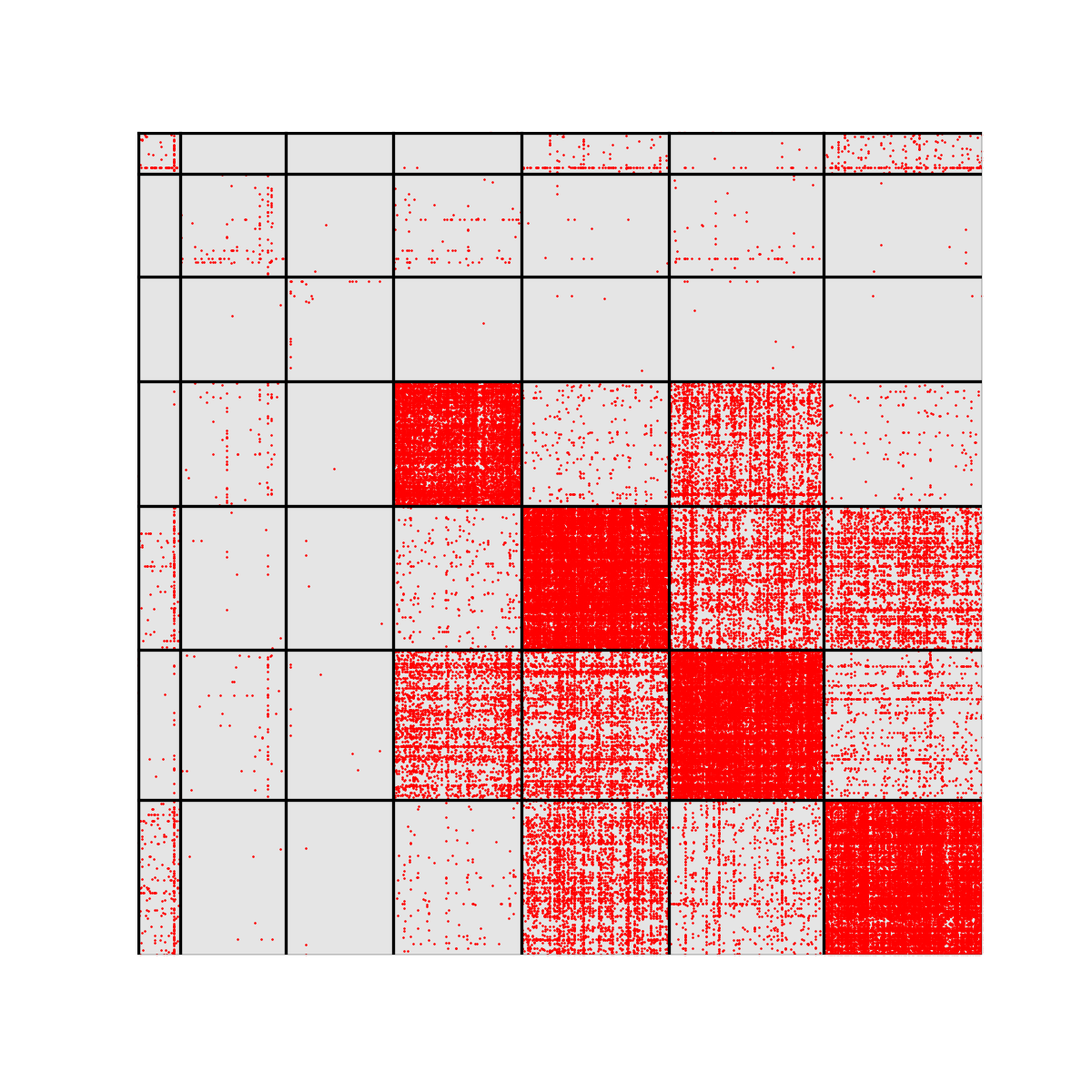}
    \caption{\scriptsize\textsc{Negative Adjacency}}
    \label{fig:dataset_pol_3}
\end{subfigure}
\hfill
\begin{subfigure}[t]{0.23\textwidth}  
    \centering
    \includegraphics[width=1\linewidth]{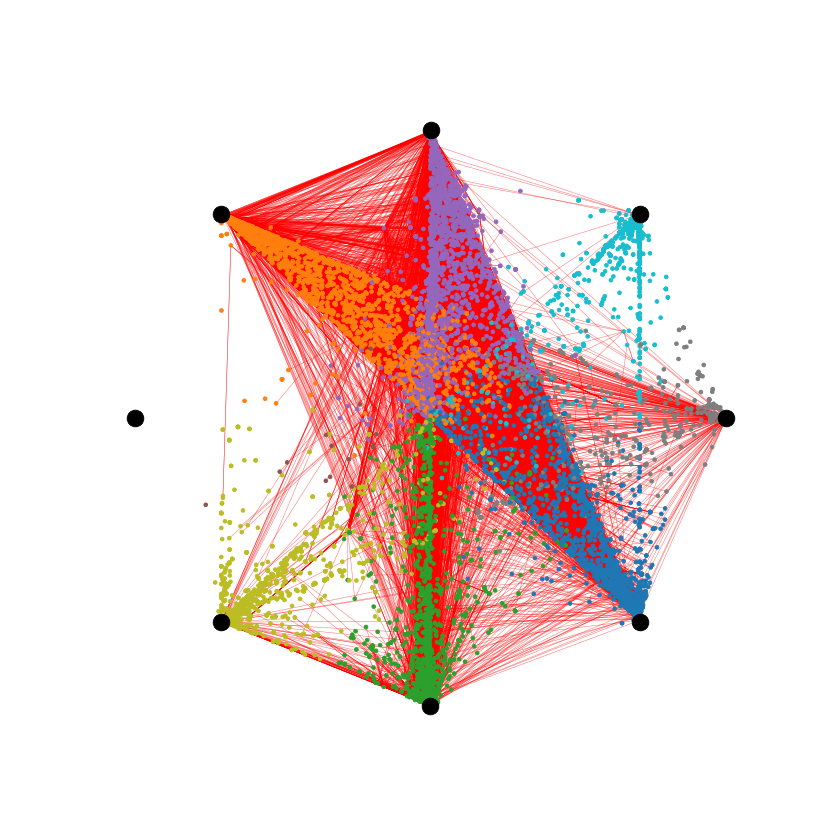}
    \caption{\scriptsize\textsc{Negative membership space}}
    \label{fig:dataset_pol_4}
\end{subfigure}
\caption{\textsc{Network Visualizations:} for the \textsl{WikiRfa} where we show the re-ordered adjacency matrix based on the maximum positive/negative memberships to the archetypes---the positive/negative membership space where essentially we visualize the soft-memberships over the archetypes in a circular plot where each archetype is positioned every $\frac{2\pi}{K}$ rads. }
\label{fig:data_wik_rfa}
\end{figure}

\textbf{Characterizing both polarization levels.} We consider the two polarized networks of Figure \ref{fig:2pols} and compare our \textsc{SGAAE} with \textsc{SLIM} in terms of expressing the two levels of network polarization. We use this baseline as it is the only direct competitor defining polarization memberships. Results are provided in Figure \ref{fig:models_2pols}, where we present the re-ordered adjacency matrices based on the structures uncovered by the two models. We witness how \textsc{SLIM} successfully characterizes the \textsc{1-level} polarization but fails in the more advanced task of the \textsc{2-level} polarization. On the other hand, \textsc{SGAAE} successfully infers all latent structures, as powered by disentangling community memberships in terms of both positive and negative structures.

\textbf{Link prediction.} We assess the performance of the proposed method in the task of sign prediction and signed link prediction, to quantify its capacity to infer meaningful links between nodes along with the sign, \ie positive or negative, of these links. In this experiment, we eliminate $20\%$ of network links and learn node embeddings based on the residual network. To create zero instances in the test set, the eliminated edges are coupled with a sample of the same number of node pairs, not showing as edges in the original network. Link signs in signed graphs introduce additional complexity to various tasks, leading to different types of prediction challenges. We address the tasks of (1) \textit{Link Sign Prediction} (p@n), and (2) \textit{Signed Link Prediction} (p@z, n@z), following the experimental setup as in \cite{slf,huang2022pole,SLIM}. Specifically, the p@n task involves predicting the sign of a removed link, assessing the model’s ability to differentiate between positive and negative links. The p@z task involves positive versus zero link prediction, assessing the model’s ability to predict the presence of positive links. The n@z task involves negative versus zero link prediction, focusing on predicting the presence of negative links. We use robust assessment measures such as the precision-recall (AUC-PR) curve (AUC-ROC scores are provided in Table 2 in the supplementary), as a result of the sparsity and class imbalances found in signed networks. 

For the \textit{link sign prediction} task, denoted as $p@n$, the AUC-PR values for the undirected case are shown in Table \ref{tab:auc_pr}. We observe that the proposed \textsc{SGAAE} model lies among the best-performing methods, while in a few cases is slightly outperformed by \textsc{SLF}. In the more challenging \textit{signed link prediction} task, denoted as $p@z$ for positive and as $n@z$ for negative samples, we predict deleted links against disconnected network pairs and accurately determine each link's sign. The test set is divided into the positive and negative disconnected subsets, and models performance is assessed on those subgroups. AUC-PR scores are shown in Table \ref{tab:auc_pr}. The proposed \textsc{SGAAE} method outcompetes all baselines for almost all datasets. We can observe that the performance improvements are, in several cases, significant, \eg scores on the \textsl{WikiElec} dataset. Here, we highlight the superiority in the performance, especially for the arduous task $n@z$, where we observe the benefits of decoupling the archetypal positive and negative memberships. (For the effect of dimensionality $K$ on model performance see Figure 4 in the supplementary.)

\textbf{Network visualization.} We show how our model can successfully infer archetypal structures that can characterize the negative and positive link patterns in the network. In Figure \ref{fig:data_wik_rfa}, we provide the re-ordered adjacency matrices focused on the positive and negative ties that validate the successful characterization over latent communities in the \textsc{WikiRfa} network, showcasing \textsc{2-level} polarization. Furthermore, we provide the positive/negative membership spaces and visualize the representation of each node in terms of the network archetypes. We observe that some archetypes are solely populated under one of the membership matrices $\mathbf{Z}$ or $\mathbf{W}$, yielding archetypes that describe groups formed uniquely under the positive or negative link structure. Consequently, this verifies the importance of accounting for different levels of polarization.

The novelty and performance gain of our model lie in its \textsc{2-level} polarization characterization, which disentangles the latent spaces into positive and negative components. Our results confirm that the most challenging task in signed link prediction is distinguishing negative links from non-links, and our model significantly outperforms baselines on the \textsl{WikiElec} and \textsl{WikiRfa} datasets -- achieving an average improvement of 30\% (11\% over the second-best) on \textsl{WikiElec} and 24\% (5\% over the most competitive) on \textsl{WikiRfa} in the n@z task. Figure \ref{fig:data_wik_rfa} illustrates how these networks exhibit \textsc{2-level} polarization by uncovering distinct communities based on positive and negative structures, leading to a more precise negative link characterization. In contrast, the Twitter network (Figure 6 in the supplementary) reflects a one-level polarization scenario, where our model matches the performance of the one-level polarization model SLIM -- a finding that corroborates our artificial network experiments in Figure \ref{fig:models_2pols}. For the \textsl{Reddit} dataset, characterized by a strong "us-versus-them" dynamic, our model outperforms \textsc{SLIM} by 1.6\% in the n@z task. Overall, our model effectively captures two-level polarization where it exists and performs comparably to one-level models when it does not, highlighting its robust performance.

%% file: 5-conclusion.tex
\section{Conclusion, Limitations \& Impact}
In this paper, we presented \textsc{SGAAE}, a signed graph AE that extracts node-level representations that express node memberships over distinct extreme profiles. This is achieved by projecting the graph onto a learned polytope, which allows for polarization characterization. We showcased how our model can account for different levels of polarization, coupled with state-of-the-art performance in link prediction for signed networks. Our concept of \textsc{2-level} polarization is in agreement with recent works providing real-world evidence that \textsc{1-level} polarization fails to capture the complexity of actual networks \citep{doi:10.1073/pnas.2401257121}. Specifically, the authors. observed that in multiple village social networks, the majority of negative ties occur within communities rather than between them. This finding suggests that the structure of negative links can develop independently from the positive network structure -- a departure from traditional approaches in signed network analysis that have predominantly focused on inter-community dynamics. In terms of limitations, \textsc{SGAAE} defines a high number of model parameters making the optimization highly non-convex and prone to local minima. Understanding polarization in social networks holds significant broader and societal impacts. Our proposed \textsc{SGAAE} framework can offer insights into the mechanisms driving division and echo chambers within online communities, as enabled by the archetypal social space.